\documentclass[10pt,twocolumn,letterpaper]{article}

\usepackage{iccv}
\usepackage{times}
\usepackage{epsfig}
\usepackage{graphicx}
\usepackage{amsmath}
\usepackage{amssymb}
\usepackage{booktabs}
\usepackage{multirow}
\usepackage{lipsum}

\usepackage[pagebackref=true,breaklinks=true,letterpaper=true,colorlinks,bookmarks=false]{hyperref} 

\usepackage[capitalize]{cleveref}
\usepackage[font={small}]{caption}
\usepackage{float}

\crefname{section}{Sec.}{Secs.}
\Crefname{section}{Section}{Sections}
\Crefname{table}{Table}{Tables}
\crefname{table}{Tab.}{Tabs.}

\newcommand\blfootnote[1]{%
    \begingroup
    \renewcommand\thefootnote{}\footnote{#1}%
    \addtocounter{footnote}{-1}%
    \endgroup
}

\DeclareMathOperator*{\argmax}{argmax}

\def\bert{BERT\xspace}
\def\vbert{VideoBERT\xspace}

\iccvfinalcopy % *** Uncomment this line for the final submission

\def\iccvPaperID{8756} % *** Enter the ICCV Paper ID here

\begin{document}

%%%%%%%%% TITLE
\title{Learning and Verification of Task Structure in Instructional Videos}

\author{Medhini Narasimhan$^{1,2}$, Licheng Yu$^{2}$,
Sean Bell$^{2}$, Ning Zhang$^{2}$, Trevor Darrell$^{1}$\vspace*{0.1in}\\
$^1$UC Berkeley, $^2$Meta AI \\
\url{https://medhini.github.io/task_structure}
}

\maketitle
\ificcvfinal\thispagestyle{empty}\fi

%%%%%%%%% ABSTRACT
\begin{abstract}
Given the enormous number of instructional videos available online, learning a diverse array of multi-step task models from videos is an appealing goal. We introduce a new pre-trained video model, VideoTaskformer, focused on representing the semantics and structure of instructional videos. We pre-train VideoTaskformer using a simple and effective objective: predicting weakly supervised textual labels for steps that are randomly masked out from an instructional video (masked step modeling). Compared to prior work which learns step representations locally, our approach involves learning them globally, leveraging video of the entire surrounding task as context. From these learned representations, we can verify if an unseen video correctly executes a given task, as well as forecast which steps are likely to be taken after a given step. \blfootnote{$^*$Work done while an intern at Meta AI. Correspondence to \texttt{medhini@berkeley.edu}}  
We introduce two new benchmarks for detecting mistakes in instructional videos, to verify if there is an anomalous step and if steps are executed in the right order. We also introduce a long-term forecasting benchmark, where the goal is to predict long-range future steps from a given step. Our method outperforms previous baselines on these tasks, and we believe the tasks will be a valuable way for the community to measure the quality of step representations.  Additionally, we evaluate VideoTaskformer on 3 existing benchmarks---procedural activity recognition, step classification, and step forecasting---and demonstrate on each that our method outperforms existing baselines and achieves new state-of-the-art performance.
\end{abstract}

\section{Introduction}
\begin{figure}[t]
    \centering
    \includegraphics[width=0.48\textwidth]{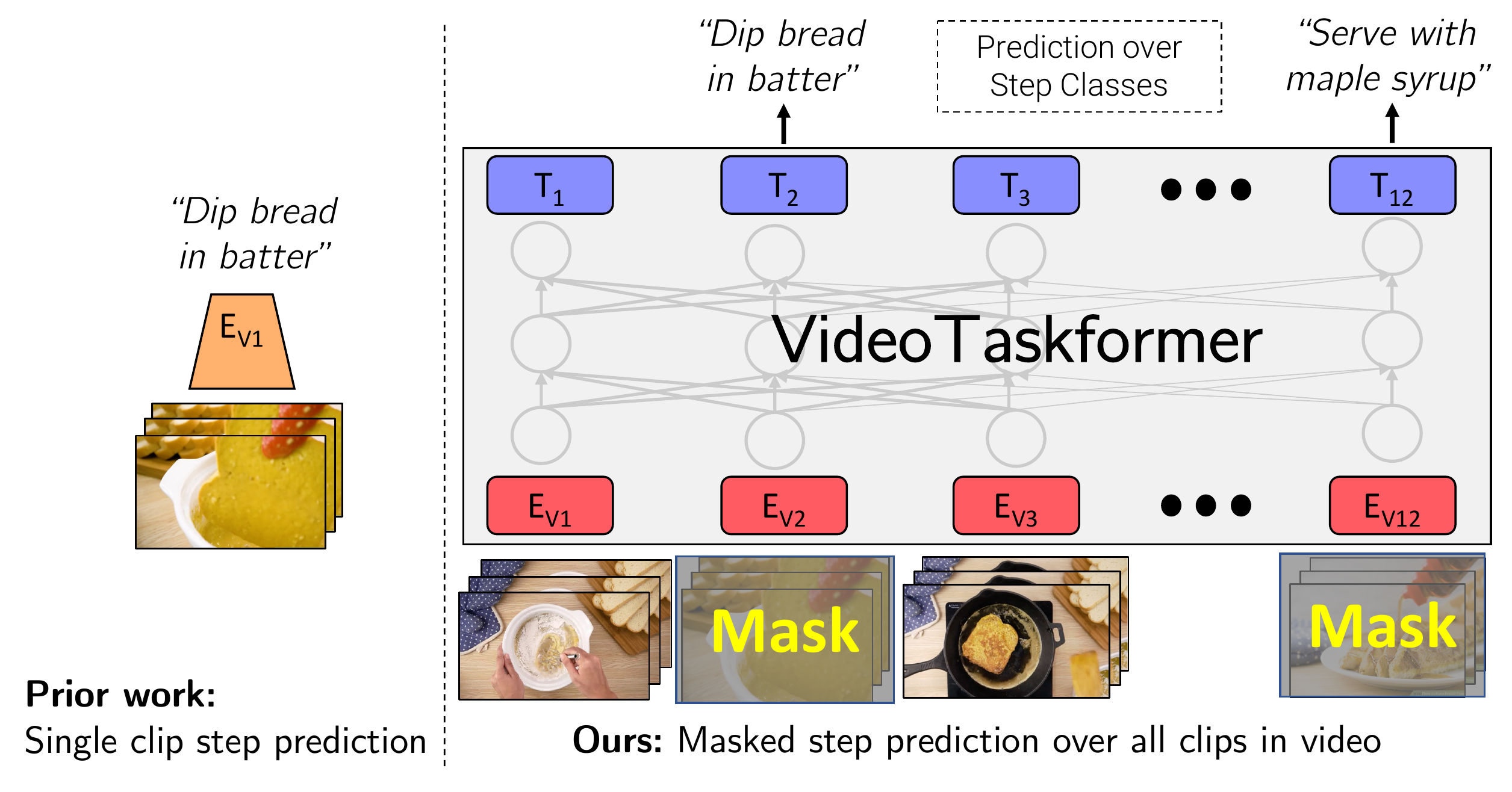}
    \caption{Prior work~\cite{lin2022learning,lei2021less} learns step representations from single short video clips, independent of the task, thus lacking knowledge of task structure. Our model, VideoTaskformer, learns step representations for masked video steps through the global context of all surrounding steps in the video, making our learned representations aware of task semantics and structure.}
    \label{fig:overview}
\end{figure}

Picture this, you’re trying to build a bookshelf by watching a YouTube video with several intricate steps. You’re annoyed by the need to repeatedly hit pause on the video and you’re unsure if you have gotten all the steps right so far. 
Fortunately, you have an interactive assistant that can guide you through the task at your own pace, verifying each step as you perform it and interrupting you if you make a mistake. 
A composite task such as ``\emph{making a bookshelf}'' involves multiple fine-grained activities such as ``\emph{drilling holes}'' and ``\emph{adding support blocks}.'' Accurately categorizing these activities requires not only recognizing the individual steps that compose the task but also understanding the task structure, which includes the temporal ordering of the steps and multiple plausible ways of executing a step (e.g., one can beat eggs with a fork or a whisk). 
An ideal interactive assistant has both a high-level understanding of a broad range of tasks, as well as a low-level understanding of the intricate steps in the tasks, their temporal ordering, and the multiple ways of performing them.

As seen in Fig.~\ref{fig:overview}, prior work~\cite{lei2021less,lin2022learning} models  step representations of a single step independent of the overall task context. 
This might not be the best strategy, given that steps for a task are related, and the way a step is situated in an overall task may contain important information about the step. To address this, we pre-train our model with a masked modeling objective that encourages the step representations to capture the \emph{global context} of the entire video. Prior work lacks a benchmark for detecting mistakes in videos, which is a crucial component of verifying the quality of instructional video representations. We introduce a mistake detection task and dataset for verifying if the task in a video is executed correctly---i.e. if each step is executed correctly and in the right order.    

% Our goal is to learn representations for the steps in the instructional video which capture semantics of the task being performed such that each step representation contains information about the surrounding context (other steps in the task). To achieve this, we train a video model VideoTaskformer using a BERT style masked modeling objective which predicts masked out steps in the input video.

Our goal is to learn representations for the steps in the instructional video which capture semantics of the task being performed such that each step representation contains information about the surrounding context (other steps in the task). To this end, we train a model VideoTaskformer, using a masked step pre-training approach for learning step representations in instructional videos. We learn step representations jointly for a whole video, by feeding multiple steps to a transformer, and masking out a subset. The network learns to predict labels for the masked steps given just the visual representations of the remaining steps. The learned contextual representations improve performance on downstream tasks such as forecasting steps, classifying steps, and recognizing procedures. 

% \trevor{Intro to this para was too chatty.  I'm going to rewrite to make slighly more formal...hope that's ok} 
Our approach of modeling steps further enables a new method for mistake identification. Recall, our original goal was to assist a user following an instructional video. We synthetically generate a mistakes dataset for evaluation using the step annotations in COIN~\cite{tang2019coin}. We consider two mistake types: mistakes in the steps of a task, and mistakes in the ordering of the steps of a task. For the first, we randomly replace the steps in a video with steps from a similar video. For the second, we re-order the steps in a task. We show that our network is capable of detecting both mistake types and outperforms prior methods on these tasks. 

Additionally, we evaluate representations learned by VideoTaskformer on three existing benchmarks: step classification, step forecasting, and procedural activity recognition on the COIN dataset. Our experiments show that learning step representation through masking pre-training objectives improves the performance on the downstream tasks. We will release code, models, and the mistake detection dataset and benchmark to the community.
%\trevor{also reference code?}

\section{Related Works}

\noindent \textbf{Instructional Video Datasets and Tasks.} Large-scale narrated instructional video datasets~\cite{damen2018scaling,miech2019howto100m,tang2019coin,zhou2018towards,zhukov2019cross} have paved the way for learning joint video-language representations and task structure from videos. More recently, datasets such as Assembly-101 dataset~\cite{Sener_2022_CVPR} and Ikea ASM~\cite{ben2021ikea} provide videos of people assembling and disassembling toys and furniture. Assembly-101 also contains annotations for detecting mistakes in the video. Some existing benchmarks for evaluating representations learned on instructional video datasets include step localization in videos~\cite{damen2018scaling,tang2019coin}, step classification~\cite{damen2018scaling,tang2019coin,zhukov2019cross}, procedural activity recognition~\cite{tang2019coin}, and step forecasting~\cite{lin2022learning}. In our work, we focus on a broad range of instructional videos found in HowTo100M~\cite{miech2019howto100m} and evaluate the learned representations on the downstream tasks in COIN~\cite{tang2019coin} dataset. We additionally introduce 3 new benchmarks for detecting mistakes in instructional videos and forecasting long-term activities. 

\noindent \textbf{Procedure Learning from Instructional Videos.} Recent works have attempted to learn procedures from instructional videos~\cite{bansal2022my, chang2020procedure, lin2022learning, qian2022svip, wang2022multimedia}. Most notably, \cite{chang2020procedure} generates a sequence of actions given a start and a goal image. \cite{bansal2022my} finds temporal correspondences between key steps across multiple videos while \cite{qian2022svip} distinguishes pairs of videos performing the same sequence of actions from negative ones. \cite{lin2022learning} uses distant supervision from WikiHow to localize steps in instructional videos. Contrary to prior works, our step representations are aware of the task structure as we learn representations globally for all steps in a video jointly, as opposed to locally, as done in past works. 

\noindent \textbf{Video Representation Learning.} There has been significant improvement in video action recognition models over the last few years \cite{arnab2021vivit,fan2021multiscale,feichtenhofer2019slowfast,liu2022video}. All of the above methods look at trimmed videos and focus on learning short-range atomic actions. In this work, we build a model that can learn longer and more complex actions, or steps, composed of multiple short-range actions. For example, the first step in Fig.~\ref{fig:overview}, \emph{``Make batter''}, is composed of several atomic actions such as \emph{``pour flour''} and \emph{``whisk''}. There have also been works~\cite{lin2022learning,miech2020end,qiu2017learning,sun2019videobert,xu2015discriminative} which learn representations for longer video clips containing semantically more complex actions. Our work falls into this line of work.

\section{Learning Task Structure through Masked Modeling of Steps}
\label{sec:method}
Our goal is to learn task-aware step representations from a large corpus of instructional videos. To this end, we develop VideoTaskformer, a video model pre-trained using a \bert~\cite{devlin2019bert} style masked modeling loss. In contrast to \bert and \vbert~\cite{sun2019videobert}, we perform masking at the step level, which encourages the network to learn step embeddings that encapsulate the semantics and temporal ordering of steps within the task. 

Our framework consists of two steps: pre-training and fine-tuning. During pre-training, VideoTaskformer is trained on weakly labeled data on the pre-training task. For fine-tuning, VideoTaskformer is first initialized with the pre-trained parameters, and a subset of the parameters is fine-tuned using labeled data from the downstream tasks. Each downstream task yields a separate fine-tuned model.

We first provide an overview of the pre-training approach before delving into details of the individual components.

\begin{figure*}[t]
    \centering
    \includegraphics[width=0.9\textwidth]{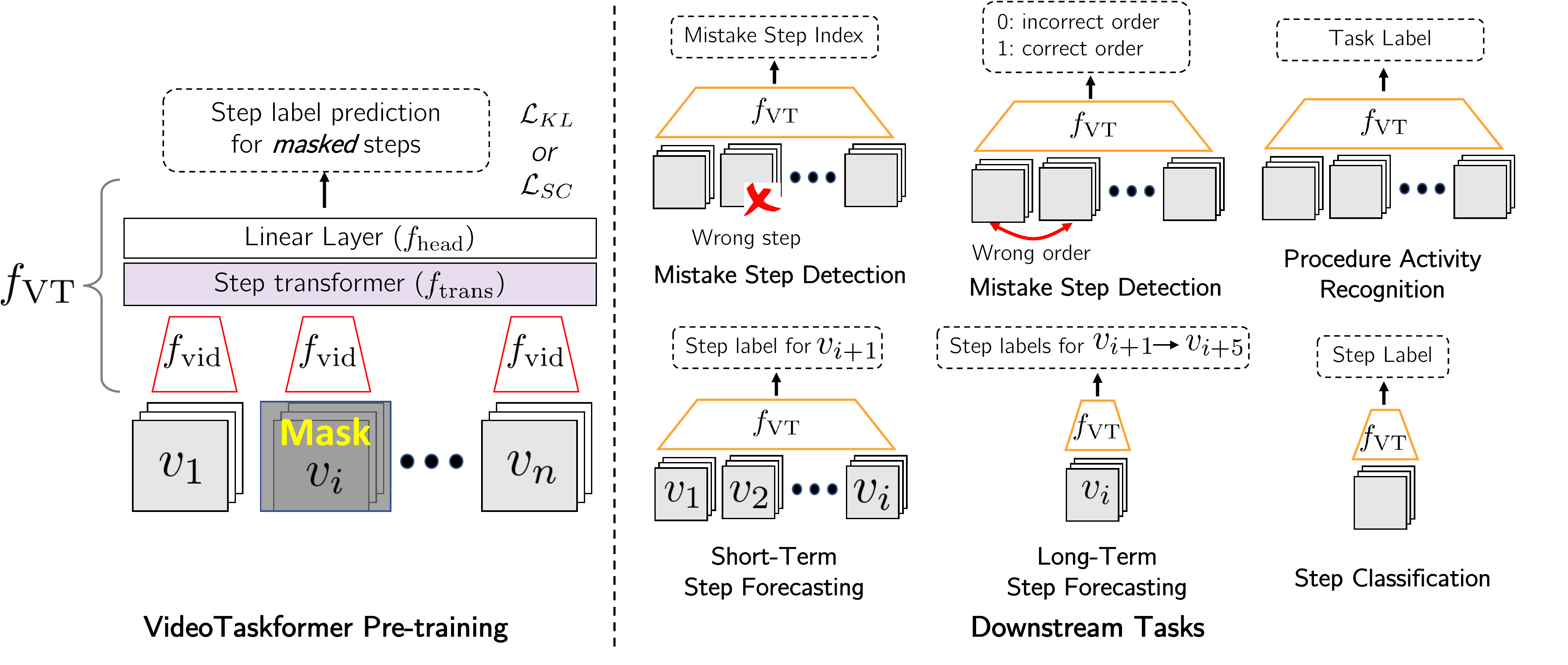}
    \caption{\textbf{VideoTaskformer Pre-training~(Left).} VideoTaskformer $f_{\text{VT}}$ learns step representations for the masked out video clip $v_i$, while attending to the other clips in the video. It consists of a video encoder $f_{\text{vid}}$, a step transformer $f_{\text{trans}}$, and a linear layer $f_{\text{head}}$, and is trained using weakly supervised step labels. \textbf{Downstream Tasks~(Right).} We evaluate step representations learned from VideoTaskformer on 6 downstream tasks. 
    }
    \label{fig:model}
\end{figure*}

\noindent \textbf{Overview.} Our approach for pre-training VideoTaskformer is outlined in Fig.~\ref{fig:model}. Consider an instructional video $V$ consisting of $K$ video clips $v_{i}, i \in [1, \dots, K]$ corresponding to $K$ steps in the video. A step $v_{i} \in \mathbb{R}^{L \times H \times W \times 3}$ is a sequence of $L$ consecutive frames depicting a step, or semantic component of the task. For example, for the task ``\emph{Making a french toast}'', examples of steps include ``\emph{Whisk the batter}'', and ``\emph{Dip bread in batter}.'' We train a video model VideoTaskformer $f_{\text{VT}}$ to learn step representations. We mask out a few clips in the input $V$ and feed it to $f_{\text{VT}}$ which learns to predict step labels for the masked-out clips. We evaluate the embeddings learned by our pre-training objective on 6 downstream tasks: step classification, procedural activity recognition, step forecasting, mistake step detection, mistake ordering detection, and long term forecasting.

% We encode the video frames for a step using the Timesformer model and feed all of its output step embeddings to a step transformer, while masking out a few steps. The Transformer predicts the step label for all the masked-out steps. This way, the network is encouraged to learn context-aware step embeddings for all steps across all videos.  

Below, we provide more details on how we pre-train VideoTaskformer using a masked step modeling loss, followed by fine-tuning details on the downstream tasks.

\subsection{Pre-training VideoTaskformer with Masked Step Modeling} 
We extend masked language modeling techniques used in \bert and \vbert to learn step representations for instructional videos. While \bert and \vbert operate on language and visual tokens respectively, VideoTaskformer operates on clips corresponding to steps in an instructional video. By predicting weakly supervised natural language step labels for masked out clips in the input video, VideoTaskformer learns semantics and long-range temporal interactions between the steps in a task. Unlike prior works wherein step representations are learned from local short video snippets corresponding to the step, our step representations are from the entire video with all the steps as input and capture \emph{global context} of the video.

\noindent \textbf{Masked Step Modeling.}
Let $V = \{v_1, \dots, v_K\}$ denote the visual clips corresponding to $K$ steps in video $V$. The goal of our our Masked Step Modeling pre-training setup is to encourage VideoTaskformer to learn representations of clips $v_i$ that are aware of the semantics of the corresponding step and the context of the surrounding task. To this end, the task for pre-training is to predict categorical natural language step labels for the masked out steps. While we do not have ground truth step labels, we use the weak supervision procedure proposed by \cite{lin2022learning} to map each clip $v_i$ to a distribution over step labels $p(y_i \mid v_i)$ by leveraging the noisy ASR annotations associated with each clip. The distribution $p(y_i \mid v_i)$ is a categorical distribution over a finite set of step labels $Y$. More details are provided in Sec.~\ref{sec:impl}.

Let $M \subseteq [1, \ldots, K]$ denote some subset of clip indices (where each index is included in $M$ with some masking probability $r$, a hyperparameter). Let $V_{\setminus M}$ denote a partially masked-out sequence of clips: the same sequence as $V$ except with clips $v_i$ masked out for all $i \in M$.

Let $f_{\text{VT}}$ represent our VideoTaskformer model with parameters $\theta$. $f_{\text{VT}}$ is composed of a video encoder model $f_{\text{vid}}$ which encodes each clip $v_i$ independently, followed by a step transformer $f_{\text{trans}}$ operating over the sequence of clip representations, and finally a linear layer $f_{\text{head}}$ (which includes a softmax). The input to the model is an entire video (of size $K \times L \times H \times W \times 3$) and the output is of size $K \times S$ (where $S$ is the output dimension of the linear layer).

We pre-train $f_{\text{VT}}$ by inputting a masked video $V_{\setminus M}$ and predicting step labels $y_{i}$ for each masked-out clip $v_i$, as described below. For the downstream tasks, we extract step-aware representations using $f_{\text{VT}}$ by feeding an unmasked video $V$ to the model. We then extract the intermediate outputs of $f_{\text{trans}}$ (which are of size $K \times D$, where $D$ is the output embedding size).

% \noindent \textbf{Learning.} To train our model $f_{\text{VT}}$, we require step labels for every video clip $v_i$ in the input video. We describe how we obtain these step labels and the approximate target distribution of video clips over step labels $P(y_s|v_i)$, in Sec.~\ref{sec:impl}. We consider the two different training objectives to train the network: (1) step classification and (2) distribution matching. Our choice of these objectives is inspired by Lin \emph{et al.}~\cite{lin2022learning}, and we describe them below for the context of masked step modeling.

To predict step labels for masked-out steps at pre-training time, we consider two training objectives: (1) step classification, and (2) distribution matching. We describe them below in the context of Masked Step Modeling.

% \noindent \textbf{Step classification loss.}
% We train $f_{\text{VT}}(\xxl) \to [0,1]^S$ to classify each \emph{masked out} video segment in the input into one of the $S$ possible steps, where $S = \{s_1, \ldots, s_N$\}. 
% Let $n^*,s^*$ be the index of the step (describing a timestep and step index) that best describes segment $v_i$ according to our target distribution $P(y_s|v_i)$. 
% We use cross-entropy loss to train $f_{\text{VT}}$ to classify video segment $v_i$ into class $(n^*,s^*)$:
% \begin{equation}
%     \min _\theta~-\log\left( \left[ f_{\text{VT}}(\xxl; \theta)  \right]_{(n^*,s^*)} \right)
% \end{equation}
% where $\theta$ denotes the learning parameters of the video model. The model uses a softmax activation function in the last layer to define a proper distribution over the steps, such that $\sum_{n,s} \left[ f_{\text{VT}}(\xxl; \theta)  \right]_{(n,s)}  = 1$. The loss above is shown for a single masked segment. However, during training, we mask 15\% of the $K$ segments in a video and optimize the objective by averaging over a mini-batch of video segments sampled from the entire collection in each iteration~\cite{lin2022learning}.

\noindent \textbf{Step classification loss.}
We use the outputs of $f_{\text{VT}}$ to represent an $S$-dimensional prediction distribution over steps, where $S=|Y|$. We form the target distribution by placing all probability mass on the best textual step description $y_i^*$ for each clip $v_i$ according to the weak supervision process. That is, 
\begin{equation}
    y_i^* = \argmax_{y \in Y} p(y \mid v_i).
\end{equation}
We calculate the cross entropy between the predicted and target distributions for each masked out clip, yielding the following expression:
\begin{equation}
    -\log([f_{\text{VT}}(V_{\setminus M})]_{j})
\end{equation}
where $j$ is the index of $y_i^*$ in $Y$, i.e., such that $y_i^* = Y_j$. To get the final training objective for a single masked video $V_{\setminus M}$, we sum over all indices $i \in M$, and minimize with respect to $\theta$.

\noindent \textbf{Distribution matching loss.} For this objective, we treat the distribution of step labels $p(y_i \mid v_i)$ from weak supervision as the target distribution for each clip $v_i$. We then compute the KL Divergence between the prediction distribution $f_{\text{VT}}(V_{\setminus M})$ and the target distribution $p(y_i \mid v_i)$ as follows:
\begin{equation}
    \sum_{j'=1}^S p(Y_{j'} \mid v_i) \log{\frac{p(Y_{j'} \mid v_i)}{[f_{\text{VT}}(V_{\setminus M})]_{j'}}}
\end{equation}
We sum over all $i \in M$ and minimize with respect to $\theta$.
Following \cite{lin2022learning}, we use only the top-$k$ steps in $p(y_i \mid v_i)$ and set the probability of the remaining steps to 0.

%We also train the step classification model $f_{\text{VT}}(\xxl)$ to minimize the KL-Divergence between the predicted distribution $f_{\text{VT}}(\xxl)$ and the target distribution $P(y_s|v_i)$ in Eq.~\ref{eq:asr_step}:
% \begin{equation} 
%     \min _\theta \sum_{n,s} P(y_s|v_i) \log \frac{P(y_s|v_i)}{\left[ f_{\text{VT}}(\xxl; \theta)  \right]_{(n,s)}}.
% \end{equation}

Lin \emph{et al.}~\cite{lin2022learning} show that the distribution matching loss results in a slight improvement over step classification loss. For VideoTaskformer, we find both objectives to have similar performance and step classification outperforms distribution matching on some downstream tasks. 

% The final equations for VideoTaskformer are:

% \begin{align}
%     \label{eq:ivsum}
%     y'_{v_i} &= f_{\text{VT}}(\xxl)\\
%     &= f_{\text{head}}(f_{\text{trans}}(\text{concat}(f_{\text{vid}}(\xxl)))) \; \forall \; i \in K\\
%     \mathcal{L}_{KL} &= \sum_{i \in \mathcal{N}} \text{KLDivergence}\: (\left[ f_{\text{VT}}(\xxl; \theta)  \right],P(y_s|v_i) \\
%     \mathcal{L}_{SC} &= \sum_{i \in \mathcal{N}} \text{CrossEntropy}\: (y_{v_i}; y'_{v_i})
%     \label{eq:mse}
% \end{align}

We use $f_{\text{VT}}$ as a feature extractor (layer before softmax) to extract step representations for new video segments. 

\subsection{Downstream Tasks}
\label{sec:tasks}

To show that the step representations learned by VideoTaskformer capture task structure and semantics, we evaluate the representations on 6 downstream tasks---3 new tasks which we introduce (mistake step detection, mistake ordering detection, and long-term step forecasting) and 3 existing benchmarks (step classification, procedural activity recognition, and short-term step forecasting). We describe the dataset creation details for our 3 new benchmarks in Sec.~\ref{sec:dataset}.

% What is a good way to determine if a video model can fully comprehend the semantics of a task from an instructional video? Here we propose one way of evaluating this - checking if the video model can detect irregularities in an instructional video. \ning{I dislike using } We introduce and evaluate our method on two new benchmarks, Mistake step detection and Mistake ordering detection.
% In this work, we introduce a new evaluation benchmark for detecting incorrect steps in videos. We propose two simple strategies to synthetically introduce mistakes in videos. Given videos from the COIN dataset where we know the start and end of each step, we construct the following datasets:
% (i) Mistake order detection: randomly re-order the steps to generate a new video where the steps are in the incorrect order
% (ii) Mistake step detection: replace a randomly chosen step in the video with a step from another video

\noindent\textbf{Mistake Detection.} A critical aspect of step representations that are successful at capturing the semantics and structure of a task is that, from these representations, \textit{correctness} of task execution can be verified. We consider two axes of correctness: content (what steps are portrayed in the video) and ordering (how the steps are temporally ordered). We introduce 2 new benchmark tasks to test these aspects of correctness.

\noindent\textbullet~\textbf{Mistake step detection.} 
The goal of this task is to identify which step in a video is incorrect. More specifically, each input consists of a video $V = \{v_{1}, \dots, v_{K}\}$ with $K$ steps.  $V$ is identical to some unaltered video $V_1$ that demonstrates a correctly executed task, except that step $v_j$ (for some randomly selected $j \in [1, \dots, K]$) is replaced with a random step from a different video $V_2$. The goal of the task is to predict the index $j$ of the incorrect step in the video.

\noindent\textbullet~\textbf{Mistake ordering detection.}
In this task, the goal is to verify if the steps in a video are in the correct temporal order. The input consists of a video $V = \{v_{1}, \dots, v_{K}\}$ with $K$ steps. There is a 50\% probability that $V$ is identical to some (correctly ordered) video $V_1 = \{v^1_{1}, \dots, v^1_{K}\}$, and there is a 50\% probability that the steps are randomly permuted. That is, $v_i = v^1_{\pi_i}$ for some random permutation $\pi$ of indices $[1,\dots, K]$. The goal of the task is to predict whether the steps are ordered correctly or are permuted. 

\noindent\textbf{Step Forecasting.} As another way to evaluate how learned step representations capture task structure, we test the capabilities of our model in anticipating future steps given one or more clips of a video.   

\noindent\textbullet~\textbf{Short-term forecasting.}
Consider a video $V = \{v_1, \dots, v_n, v_{n+1}, \dots v_K\}$ where $v_i$ denotes a step, and $V$ has step labels $\{y_1, \dots, y_K\}$, where $y_i \in Y$, the finite set of all step labels in the dataset. Short-term forecasting involves predicting the step label $y_{n+1}$ given the previous $n$ segments $\{v_1, \dots, v_n\}$~\cite{lin2022learning}.   

\noindent\textbullet~\textbf{Long-term step forecasting.} We introduce the challenging task of long-term step forecasting. Given a single step $v_i$ in a video $V = \{v_1, \dots, v_K\}$ with step labels $\{y_1, \dots, y_K\}$, the task is to predict the step labels for the next 5 steps, i.e. $\{y_{i+1}, y_{i+2}, \ldots, y_{i+5}\}$. This task is particularly challenging since the network receives very little context---just a single step---and needs to leverage task information learned during training from watching multiple different ways of executing the same task.    

\noindent\textbf{Procedural Activity Recognition.} The goal of this task is to recognize the procedural activity (i.e., task label) from a long instructional video. The input to the network is all the $K$ video clips corresponding to the steps in a video, $V = \{v_1, \dots, v_K\}$. The task is to predict the video task label $t \in \mathcal{T}$ where $\mathcal{T}$ is the set of all task labels for all the videos in the dataset.

\noindent\textbf{Step Classification.} In this task, the goal is to predict the step label $y_i \in Y$ given the video clip corresponding to step $v_i$ from a video $V = \{v_1, \dots, v_K\}$. No context other than the single clip is given. Therefore, this task requires fine-grained recognition capability, which would benefit from representations that contain information about the context in which a step gets performed.

For all of the above tasks, we use the step and task label annotations as supervision. We show the ``zero-shot'' performance of VideoTaskformer by keeping the video model $f_{\text{vid}}$ and the transformer layer $f_{\text{trans}}$ fixed and only fine-tuning a linear head $f_{\text{head}}$ on top of the output representations. Additionally, we also show fine-tuning results where we keep the base video model $f_{\text{vid}}$ fixed and fine-tune the final transformer $f_{\text{trans}}$ and the linear layer $f_{\text{head}}$ on top of it. The network is fine-tuned using cross-entropy loss with supervision from the step labels for all downstream tasks. 

\subsection{Implementation Details}
\label{sec:impl}
\noindent \textbf{Step labels from Weak Supervision.} To train VideoTaskformer, we require step annotations, i.e., step labels with start and end timestamps in the video, for a large corpus of instructional videos. Unfortunately, this is difficult to obtain manually and datasets that provide these annotations, like COIN and CrossTask, are small in size ($\sim$10K videos). To overcome this issue,  the video speech transcript can be mapped to steps in WikiHow and used as a weak form of supervision \cite{lin2022learning}. The intuition behind this is that WikiHow steps are less noisy compared to transcribed speech.   

The WikiHow dataset contains a diverse array of articles with step-by-step instructions for performing a range of tasks. Denote the steps across all $T$ tasks in WikiHow as $s = \{s_1, \ldots, s_N$\}, where $s_n$ represents the natural language title of the $n$th step in $s$, and $N$ is the number of steps across all tasks in WikiHow. Each step $s_n$ contain a lengthy language-based description which we denote as $y_n$. 

Consider a video with $K$ sentences in the automatic speech transcript denoted as $\{a_1, \ldots, a_K\}$. Each sentence is accompanied by a $\{start, end\}$ timestamp to localize it in the video. This yields $K$ corresponding video segments denoted as $\{v_1, \ldots, v_K\}$. Each video segment $v_i$ is a sequence of $F$ RGB frames having spatial resolution $H \times W$.
To obtain the step label for a segment $v_i$,  the corresponding sentence in the transcript $a_i$ is used to find the distribution of the nearest steps in the WikiHow repository.
Following~\cite{lin2022learning}, we approximate this distribution using a textual similarity measure $\text{sim}$ between $y_n$ and $a_i$:

% \begin{equation}
%     P(y_s|v_i) \approx \frac{\exp{(\text{sim} (a_i,y_s))}}{\sum_{t,s} \exp{(\text{sim} (a_i,y_s))}}.
%     \label{eq:asr_step}
% \end{equation}

\begin{equation}
    P(y_n|v_i) \approx \frac{\exp{(\text{sim} (a_i,y_n))}}{\sum_{n'} \exp{(\text{sim} (a_i,y_{n'}))}}.
    \label{eq:asr_step}
\end{equation}

The authors of \cite{lin2022learning} found it best to search over all the steps across all tasks (i.e., all $y_n$), rather than the set of steps for the specific task referenced in the video. The  similarity function $\text{sim}$ is formulated as a dot product between language embeddings obtained from a pre-trained language model.

\noindent \textbf{Language model.} To compare WikiHow steps to the transcribed speech via the $\text{sim}$ function, we follow the same setup as in Lin \emph{et al.}~\cite{lin2022learning}. For a fair comparison to the baseline, we use MPNet (paraphrase-mpnet-base-v2)  to extract sentence embeddings $\in \mathbb{R}^{768}$. 

\noindent \textbf{Video model.} VideoTaskformer is a TimeSformer model with a two-layer transformer. Following \cite{lin2022learning}, the TimeSformer is initialized with ViT \cite{dosovitskiy2020image} pre-trained on ImageNet-21K, and is trained on subsampled clips from HowTo100M (with 8 frames sampled uniformly from each 8-second span).

We include additional implementation details in the Supplemental.

\section{Datasets and Evaluation Metrics}
\label{sec:dataset}
% insert figure from dataset 
% insert dataset statistics - number of videos (train/val/test), number of mistakes, easy/hard mistakes  

\noindent \textbf{Pre-training.} For pre-training, we use videos and transcripts from the HowTo100M (HT100M)~\cite{miech2019howto100m} dataset and steps from the WikiHow dataset~\cite{bertasius2021space}. HT100M contains 136M video clips from 1.2M long narrated instructional videos, spanning 23k activities such as ``gardening'' and ``personal care.'' The WikiHow dataset contains 10,588 steps collected from 1059 WikiHow articles which are sourced from the original dataset~\cite{koupaee2018wikihow}. 

\noindent \textbf{Evaluation.} 
All evaluation benchmarks use videos and step annotations from the COIN dataset~\cite{tang2019coin}. COIN consists of 11,827 videos related to 180 different tasks and provides step labels with start and end timestamps for every video. We use a subset of 11,104 videos that were available to download.

As described in Sec. \ref{sec:tasks}, we introduce 3 new benchmark tasks in this work: mistake step detection, mistake ordering detection, and long-term step forecasting. 

\noindent \textbf{\textit{Mistake Step Detection.}} For creating the mistake step detection dataset, for every video in the COIN dataset, we randomly replace one step with a step from a different video. The network predicts the index of the mistake step. We use the same train/validation/test splits as in COIN and report average accuracy of predicting the mistake step index on the test set. 

\noindent \textbf{\textit{Mistake Ordering Detection.}} We synthetically create the mistake ordering detection dataset by randomly shuffling the ordering of the steps in a given video, for 50\% of the videos and train the network to predict whether the steps are in the right order or not. While creating the
dataset, we repeatedly shuffle the input steps until the shuffled “mistake” order is different from the original valid order. Additionally, we compare the shuffled “mistake” order
across all the videos in the same task, to ensure it doesn’t
match any other video’s correct ordering of steps. Despite these two pre-processing checks, there might be noise in the dataset. We report average prediction accuracy on the test split.

\noindent \textbf{\textit{Long-term step forecasting.}} Given a video clip corresponding to a single step, long-term step forecasting involves predicting the step class label for the next 5 consecutive steps. If there are fewer than 5 next steps we append NULL tokens to the sequence. We compute classification accuracy as the number of correct predictions out of the total number of predictions, ignoring NULL steps. We again use the same splits in the COIN dataset. 

Additionally, we evaluate on 3 existing benchmarks: \textbf{\textit{step classification}}~\cite{tang2019coin} - predicts the step class label from a single video clip containing one step, \textbf{\textit{procedural activity recognition}}~\cite{tang2019coin} - predicts the procedure/task label given all the steps in the input video, and \textbf{\textit{short-term step forecasting}}~\cite{lin2022learning} - predicts the class of the step in the next segment given as input the sequence of observed video segments up to that step (excluded).

\section{Experiments}
\begin{table*}[h]
\centering
\footnotesize
\begin{tabular}{lllc}
\toprule 
Model   & Pre-training Supervision & Pre-training Dataset  & Acc (\%)
\\ \midrule
TSN (RGB+Flow)~\cite{tang2020comprehensive}  & Supervised: action labels & Kinetics      & 36.5* \\ 
S3D~\cite{miech2020end} & Unsupervised: MIL-NCE on ASR  & HT100M      & 37.5* \\ 
\hline
ClipBERT~\cite{lei2021less} & Supervised: captions & COCO + Visual Genome & 30.8\\
VideoCLIP~\cite{xu2021videoclip} & Unsupervised: NCE on ASR  & HT100M & 39.4 \\ 
SlowFast~\cite{feichtenhofer2019slowfast}  & Supervised: action labels  & Kinetics & 32.9 \\
TimeSformer~\cite{bertasius2021space} & Supervised: action labels & Kinetics & 48.3 \\
LwDS: TimeSformer~\cite{bertasius2021space} & Unsupervised: $k$-means on ASR & HT100M & 46.5 \\
LwDS: TimeSformer  & Distant supervision & HT100M & 54.1 \\ \hline
VideoTF (SC)  & Unsupervised: NN on ASR & HT100M & 47.0\\
\textbf{VideoTF (DM)}  & \textbf{Distant supervision} & HT100M &  \textbf{54.8} \\
\textbf{VideoTF (SC)}   & \textbf{Distant supervision} & HT100M &  \textbf{56.5} \\
\bottomrule
\end{tabular}
\caption{\textbf{Step classification.} We compare to the accuracy scores for all baselines. VideoTF (SC) pre-trained with step classification loss on distant supervision from WikiHow achieves state-of-the-art performance on the downstream step classification task. We report baseline results from \cite{lin2022learning}. * indicates results by fine-tuning on COIN}
\label{tab:step}
\end{table*}

\begin{table*}[h]
\centering
\footnotesize
\setlength{\tabcolsep}{1pt}
\begin{tabular}{lllcc}
\toprule
Downstream Model & Base Model  & Pre-training Supervision & Pre-training Dataset  & Acc (\%)
\\ \midrule
TSN (RGB+Flow)~\cite{tang2020comprehensive} & Inception~\cite{szegedy2016rethinking} & Supervised: action labels & Kinetics      & 73.4* \\ 
Transformer & S3D~\cite{miech2020end}  & Unsupervised: MIL-NCE on ASR  & HT100M     & 70.2* \\ 
Transformer & ClipBERT~\cite{lei2021less} & Supervised: captions & COCO + Visual Genome & 65.4 \\
Transformer & VideoCLIP~\cite{xu2021videoclip} & Unsupervised: NCE on ASR  & HT100M      &  72.5\\ 
Transformer & SlowFast~\cite{feichtenhofer2019slowfast}  & Supervised: action labels  & Kinetics & 71.6 \\
Transformer & TimeSformer~\cite{bertasius2021space}   & Supervised: action labels & Kinetics      & 83.5 \\
LwDS: Transformer & TimeSformer~\cite{bertasius2021space}   & Unsupervised: $k$-means on ASR & HT100M      & 85.3 \\
LwDS: Transformer & TimeSformer   & Distant supervision & HT100M     &  88.9 \\
LwDS: Transformer w/ KB Transfer & TimeSformer  & Distant supervision & HT100M     & 90.0 \\ \hline
VideoTF (SC; fine-tuning) w/ KB Transfer & TimeSformer & Unsupervised: NN on ASR & HT100M & 81.2\\
VideoTF (SC; linear-probe) w/ KB Transfer & TimeSformer & Distant supervision & HT100M & 83.1\\
VideoTF (DM; linear-probe) w/ KB Transfer & TimeSformer & Distant supervision & HT100M &  85.7\\
VideoTF (SC) w/ KB Transfer & TimeSformer & Distant supervision & HT100M & 90.5\\
\textbf{VideoTF (DM) w/ KB Transfer} & \textbf{TimeSformer} & Distant supervision & \textbf{HT100M} & \textbf{91.0} \\
\bottomrule
\end{tabular}
\centering
\caption{Accuracy of different methods on the \textbf{procedural activity recognition} dataset.}
\label{tab:proc}
\end{table*}

\begin{table*}[!htbp]
\centering
\footnotesize
\begin{tabular}{lllllll}
\toprule
Downstream Model & Base Model  & Pre-training Supervision & Pre-training Dataset  & Acc (\%)
\\ \midrule
Transformer & S3D~\cite{miech2020end}  & Unsupervised: MIL-NCE on ASR  & HT100M     & 28.1 \\ 
Transformer & SlowFast~\cite{feichtenhofer2019slowfast}  & Supervised: action labels  & Kinetics     & 25.6 \\
Transformer & TimeSformer~\cite{bertasius2021space}   & Supervised: action labels & Kinetics      & 34.7 \\
LwDS: Transformer & TimeSformer~\cite{bertasius2021space}   & Unsupervised: $k$-means on ASR & HT100M      & 34.0 \\
LwDS: Transformer w/ KB Transfer & TimeSformer  & Distant supervision & HT100M     &  39.4 \\ \hline
VideoTF (SC; fine-tuned) w/ KB Transfer & TimeSformer & Unsupervised: NN on ASR & HT100M & 35.1\\
VideoTF (SC; linear-probe) w/ KB Transfer & TimeSformer & Distant supervision & HT100M & 39.2\\
VideoTF (DM; linear-probe) w/ KB Transfer & TimeSformer & Distant supervision & HT100M &  40.1\\
VideoTF (SC) w/ KB Transfer & TimeSformer & Distant supervision & HT100M & 41.5\\
\textbf{VideoTF (DM) w/ KB Transfer} & \textbf{TimeSformer} & Distant supervision & \textbf{HT100M} & \textbf{42.4} \\
\bottomrule
\end{tabular}
\vspace{-.1cm}
\caption{Accuracy of different methods on the \textbf{short-term step forecasting}  dataset.}
\vspace{-.2cm}
\label{tab:sfore}
\end{table*}

\begin{table*}[!htbp]
\centering
\footnotesize
\begin{tabular}{lllllll}
\toprule
Downstream Model & Base Model  & Pre-training Supervision & Pre-training Dataset  & Acc (\%)
\\ \midrule
Transformer (ASR text) w/ Task label & MPNet & & & 39.0\\
Transformer & SlowFast~\cite{feichtenhofer2019slowfast}  & Supervised: action labels  & Kinetics  & 15.2\\
Transformer & TimeSformer~\cite{bertasius2021space}   & Supervised: action labels & HT100M &  17.0\\
Transformer w/ Task label & TimeSformer~\cite{bertasius2021space}   & Supervised: action labels & HT100M &  40.1\\
LwDS: Transformer w/ Task label & TimeSformer   & Distant supervision & HT100M     &  41.3 \\\hline
VideoTF (DM) & TimeSformer & Distant supervision & HT100M & 40.2\\
\textbf{VideoTF (DM) w/ Task label} & \textbf{TimeSformer} & Distant supervision & \textbf{HT100M} & \textbf{46.4} \\
\bottomrule
\end{tabular}
\vspace{-.1cm}
\caption{Accuracy of different methods on the \textbf{long-term step forecasting} dataset.}
\vspace{-.2cm}
\label{tab:lfore}
\end{table*}

\begin{table*}[!htbp]
\centering
\footnotesize
\begin{tabular}{llllcc}
\toprule
\multirow{2}{*}{Downstream Model} & \multirow{2}{*}{Base Model}  & \multirow{2}{*}{Pre-training Supervision} & \multirow{2}{*}{Pre-training Dataset}  & \multicolumn{2}{c}{Mistake Detection} \\
\multicolumn{4}{c}{} & \textbf{Step} & \textbf{Order}
\\ \midrule
Transformer (ASR text) w/~Task label & MPNet \cite{song2020mpnet} & & & 34.2 & 33.4\\
Transformer w/~Task Label & SlowFast~\cite{feichtenhofer2019slowfast}  & Supervised: action labels  & Kinetics     &  28.6 & 26.1\\
Transformer w/~Task label & TimeSformer~\cite{bertasius2021space}   & Supervised: action labels & HT100M  & 36.0 & 34.7 \\
LwDS: Transformer & TimeSformer & Distant supervision & HT100M     &  17.1 & 11.2\\ 
LwDS: Transformer w/~Task Label & TimeSformer & Distant supervision & HT100M     & 37.6 & 31.8\\ \hline
VideoTF (SC) & TimeSformer & Distant supervision & HT100M & 20.1 & 15.4\\
VideoTF (DM) w/~Task label & TimeSformer & Distant supervision & HT100M & 40.8 & 34.0\\
\textbf{VideoTF (SC; fine-tuned) w/~Task label} & \textbf{TimeSformer} & Distant supervision & \textbf{HT100M} & \textbf{41.7} & \textbf{35.4}\\
\bottomrule
\end{tabular}
\vspace{-.1cm}
\caption{Accuracy of different methods on the \textbf{mistake step detection} test dataset.}
\vspace{-.2cm}
\label{tab:mis}
\end{table*}

We evaluate VideoTaskformer~(VideoTF) and compare it with existing baselines on 6 downstream tasks: step classification, procedural activity recognition, step forecasting, mistake step detection, mistake ordering detection, and long term forecasting. Results are on the datasets described in Sec.~\ref{sec:dataset}.

\subsection{Baselines} 
We compare our method to state-of-the-art video representation learning models for action/step recognition. We fine-tune existing models in a similar fashion to ours on the 6 downstream tasks. We briefly describe the best performing baseline, Learning with Distant Supervision (LwDS)~\cite{lin2022learning}.

\noindent\textbullet~\textbf{TimeSformer~(LwDS)~\cite{lin2022learning}.} In this baseline model, the TimeSformer backbone is pre-trained on HowTo100M using the Distribution Matching loss (but without any masking of steps as in our model). Next, a single-layer transformer is fine-tuned on top of the pre-trained representations from the base model for each downstream task. 

\noindent\textbullet~\textbf{TimeSformer w/ KB transfer~(LwDS)~\cite{lin2022learning}.} For procedural activity recognition and step forecasting, the LwDS baseline is modified to include knowledge base transfer via retrieval of most relevant facts from the knowledge base to assist the downstream task. We also include results by adding the same KB transfer component to our method, referenced as  w/ KB Transfer. 

\noindent\textbullet~\textbf{Steps from clustering ASR text.} As an alternative to the weak supervision from WikiHow, we introduce an unsupervised baseline that relies only on the transcribed speech (ASR text) to obtain steps. \cite{narasimhan2022tl} introduced an approach to segment a video into steps by clustering visual features along the time axis. It divides the video into non-overlapping segments and groups adjacent video segments together based on a similarity threshold. We adopt a similar approach but in the text space. We compute sentence embeddings for the ASR sentences and group adjacent sentences if their similarity exceeds the average similarity of all sentences across the entire video. We include ablations with different thresholds in the Supplemental.     

\begin{figure*}[t]
    \centering
    \includegraphics[width=0.9\textwidth]{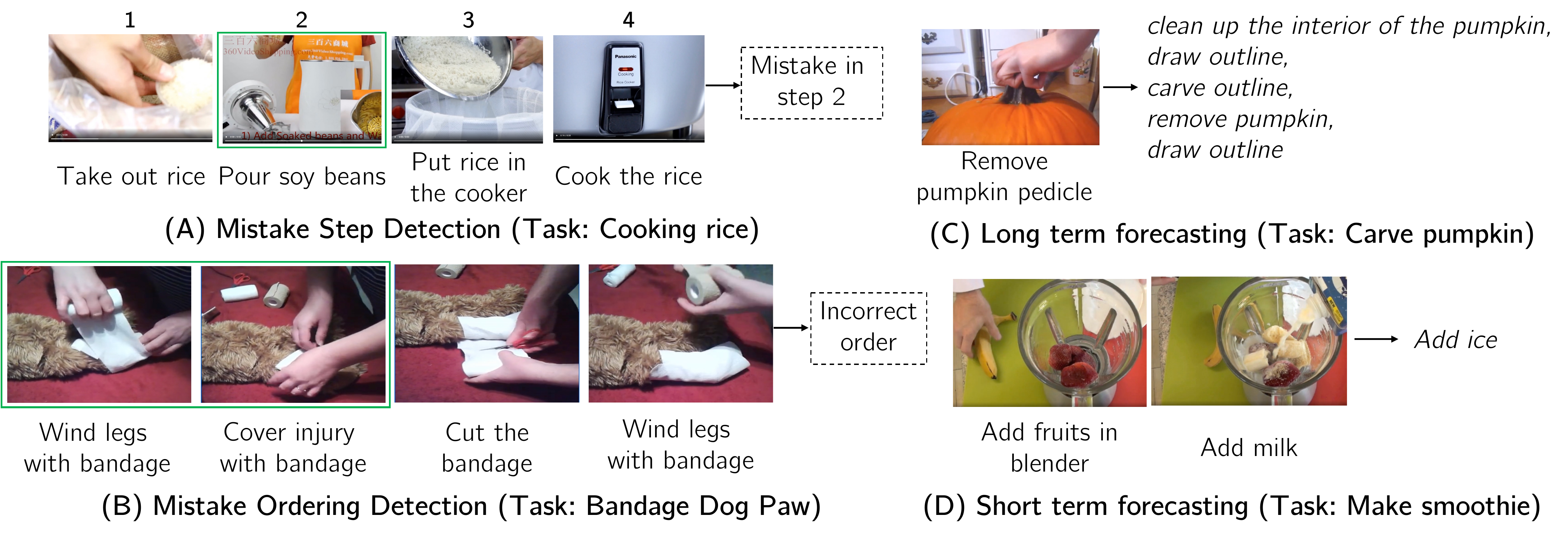}
    \caption{\textbf{Qualitative results.} We show qualitative results of our method on 4 tasks. The step labels are not used during training and are only shown here for illustrative purposes.}
    \label{fig:res}
\end{figure*}

\begin{figure}[t]
    \centering
    \includegraphics[width=0.5\textwidth]{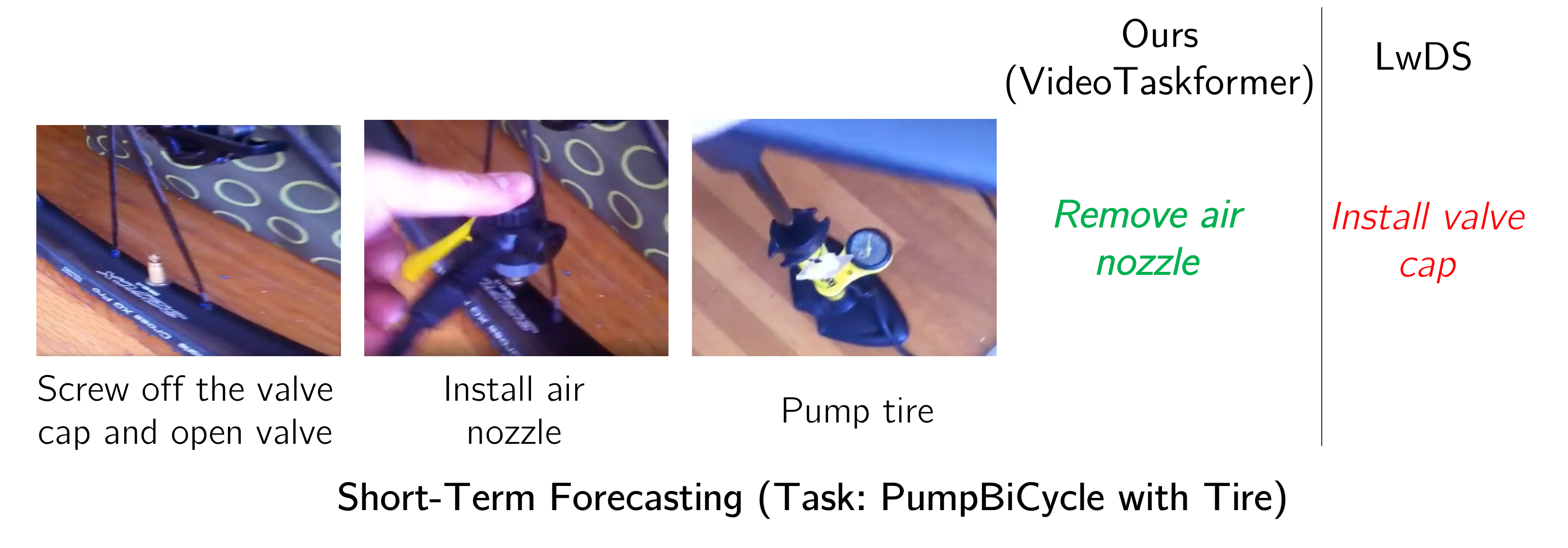}
    \caption{\textbf{Qualitative comparison.} We compare results from our method VideoTF to the baseline LwDS on the short-term forecasting task. Step labels are not passed to the model as input and are only for reference.}
    \label{fig:res2}
\end{figure}

\subsection{Ablations}
We evaluate our design choices by ablating different components of our model.  

\noindent\textbullet~\textbf{Base model.} We report results for different base video models for pre-training: S3D \cite{miech2020end}, SlowFast \cite{feichtenhofer2019slowfast}, TimeSformer \cite{bertasius2021space} trained on HT100M, and TimeSformer trained on Kinetics. For short-term step forecasting, procedural activity recognition, and step classification, the results are from \cite{lin2022learning}. 

\noindent\textbullet~\textbf{Loss function.} For pre-training VideoTF, we test both the loss functions, Step Classification (SC), and Distribution Matching (DM) described in Sec.~\ref{sec:method}.  

\noindent\textbullet~\textbf{Modalities.} For mistake step detection and long-term forecasting tasks, we tried replacing video features with ASR text during fine-tuning. The base model is a language model for embedding sentences in the ASR text and is kept fixed. The ASR text embeddings for all the segments of the video are fed as input to the downstream model, a basic single-layer transformer, which is fine-tuned to each of the tasks. 

\noindent\textbullet~\textbf{Task label.} For mistake detection and long-term forecasting tasks, we include the task name, e.g. \emph{``Install a Ceiling Fan''}, as input to the downstream model. We compute the sentence embedding of the task label and append it to the list of video tokens fed as input to the model. This domain knowledge provides additional context which boosts the performance on these challenging downstream tasks.

\noindent\textbullet~\textbf{Linear-probe vs Fine-tuning.} In linear-probe evaluation, only the $f_\text{head}$ layer is fine-tuned to each downstream task and in the fine-tuning setting, all the layers of the segment transformer $f_\text{trans}$ are fine-tuned. 

\subsection{Results}
\noindent \textbf{Quantitative Results.} We compare our approach to several baselines on all downstream tasks. For all the downstream tasks, the downstream segment transformer is fine-tuned, except for linear-probe where we keep our pre-trained model fixed and only train a linear head on top of it for each downstream task.

On the step classification task in Tab.~\ref{tab:step}, VideoTF with step classification loss outperforms LwDS~\cite{lin2022learning} by ~2\%, indicating that step representations learned with global context also transfer well to a task that only looks at local video clips. In procedural activity recognition (Tab.~\ref{tab:proc}), we see that distribution matching loss works slightly better than step classification loss and our fine-tuned model achieves 1\% improvement over the best baseline. For short-term forecasting in Tab.~\ref{tab:sfore}, we achieve a 3\% improvement over LwDS and our unsupervised pre-training using NN with ASR outperforms previous unsupervised methods. We also note that linear-probe performance is competitive in Tab.~\ref{tab:proc} and outperforms baselines in Tab.~\ref{tab:sfore}. VideoTF with achieves a strong improvement of 5\% over LwDS on the long-term forecasting task, 4\% on mistake step detection, and 4\% on mistake ordering detection. Adding task labels improves performance on all three tasks. 

Additionally, we evaluate our approach on the activity recognition task in EPIC Kitchens-100 and include results in the Supplemental. We also report our models performance on the step localization task in COIN.   

\noindent \textbf{Qualitative Results.} Fig.~\ref{fig:res} shows qualitative results of our model VideoTF on the mistake detection tasks. Fig.~\ref{fig:res} (A) shows a result on mistake step detection, where our model's input is the sequence of video clips on the left and it correctly predicts the index of the mistake step ``2'' as the output. In (B), the order of the first two steps is swapped and our model classifies the sequence as incorrectly ordered. In (C), for the long-term forecasting task, the next 5 steps predicted by our model match the ground truth and in (D), for the short-term forecasting task, the model predicts the next step correctly given the past 2 steps. In Fig.~\ref{fig:res2} we show an example result of our method compared to the baseline LwDS on the short-term forecasting  task. Our method correctly predicts the next step as ``remove air nozzle'' since it has acquired knowledge of task structure whereas the baseline predicts the next step incorrectly as ``install valve cap.''

\section{Conclusion}
In this work, we introduce a new video model, VideoTaskformer, for learning contextualized step representations through masked modeling of steps in instructional videos. 
We also introduce 3 new benchmarks: mistake step detection, mistake order detection, and long term forecasting. 
We demonstrate that VideoTaskformer improves performance on 6 downstream tasks, with particularly strong improvements in detecting mistakes in videos and long-term forecasting. 
Our method opens the possibility of learning to execute a variety of tasks by watching instructional videos; imagine learning to cook a complicated meal by watching a cooking show. \\
\textbf{Acknowledgements.} We would like to thank Suvir Mirchandani for his help with experiments and paper writing. This work was supported in part by DoD including DARPA's LwLL, PTG and/or SemaFor programs, as well as BAIR's industrial alliance programs.

%%%%%%%%% REFERENCES
{\small
\bibliographystyle{ieee_fullname}
\bibliography{egbib}

\begin{thebibliography}{10}\itemsep=-1pt

\bibitem{arnab2021vivit}
Anurag Arnab, Mostafa Dehghani, Georg Heigold, Chen Sun, Mario Lu{\v{c}}i{\'c},
  and Cordelia Schmid.
\newblock Vivit: A video vision transformer.
\newblock In {\em IEEE International Conference on Computer Vision (ICCV)},
  2021.

\bibitem{bansal2022my}
Siddhant Bansal, Chetan Arora, and CV Jawahar.
\newblock My view is the best view: Procedure learning from egocentric videos.
\newblock In {\em European Conference on Computer Vision (ECCV)}, 2022.

\bibitem{ben2021ikea}
Yizhak Ben-Shabat, Xin Yu, Fatemeh Saleh, Dylan Campbell, Cristian
  Rodriguez-Opazo, Hongdong Li, and Stephen Gould.
\newblock The ikea asm dataset: Understanding people assembling furniture
  through actions, objects and pose.
\newblock In {\em Winter Conference on Applications of Computer Vision (WACV)},
  2021.

\bibitem{bertasius2021space}
Gedas Bertasius, Heng Wang, and Lorenzo Torresani.
\newblock Is space-time attention all you need for video understanding?
\newblock In {\em International Conference on Machine Learning (ICML)}, 2021.

\bibitem{chang2020procedure}
Chien-Yi Chang, De-An Huang, Danfei Xu, Ehsan Adeli, Li Fei-Fei, and
  Juan~Carlos Niebles.
\newblock Procedure planning in instructional videos.
\newblock In {\em European Conference on Computer Vision (ECCV)}, 2020.

\bibitem{damen2018scaling}
Dima Damen, Hazel Doughty, Giovanni~Maria Farinella, Sanja Fidler, Antonino
  Furnari, Evangelos Kazakos, Davide Moltisanti, Jonathan Munro, Toby Perrett,
  Will Price, et~al.
\newblock Scaling egocentric vision: The {EPIC-KITCHENS} dataset.
\newblock In {\em European Conference on Computer Vision (ECCV)}, 2018.

\bibitem{devlin2019bert}
Jacob Devlin, Ming-Wei Chang, Kenton Lee, and Kristina Toutanova.
\newblock {BERT}: Pre-training of deep bidirectional transformers for language
  understanding.
\newblock In {\em Conference of the North American Chapter of the Association
  for Computational Linguistics}, 2019.

\bibitem{dosovitskiy2020image}
Alexey Dosovitskiy, Lucas Beyer, Alexander Kolesnikov, Dirk Weissenborn,
  Xiaohua Zhai, Thomas Unterthiner, Mostafa Dehghani, Matthias Minderer, Georg
  Heigold, Sylvain Gelly, et~al.
\newblock An image is worth 16x16 words: Transformers for image recognition at
  scale.
\newblock In {\em International Conference on Learning Representations (ICLR)},
  2020.

\bibitem{fan2021multiscale}
Haoqi Fan, Bo Xiong, Karttikeya Mangalam, Yanghao Li, Zhicheng Yan, Jitendra
  Malik, and Christoph Feichtenhofer.
\newblock Multiscale vision transformers.
\newblock In {\em IEEE International Conference on Computer Vision (ICCV)},
  2021.

\bibitem{feichtenhofer2019slowfast}
Christoph Feichtenhofer, Haoqi Fan, Jitendra Malik, and Kaiming He.
\newblock {SlowFast} networks for video recognition.
\newblock In {\em IEEE International Conference on Computer Vision (ICCV)},
  2019.

\bibitem{koupaee2018wikihow}
Mahnaz Koupaee and William~Yang Wang.
\newblock Wikihow: A large scale text summarization dataset.
\newblock {\em arXiv:1810.09305}, 2018.

\bibitem{lei2021less}
Jie Lei, Linjie Li, Luowei Zhou, Zhe Gan, Tamara~L Berg, Mohit Bansal, and
  Jingjing Liu.
\newblock Less is more: {ClipBERT} for video-and-language learning via sparse
  sampling.
\newblock In {\em IEEE Conference on Computer Vision and Pattern Recognition
  (CVPR)}, 2021.

\bibitem{lin2022learning}
Xudong Lin, Fabio Petroni, Gedas Bertasius, Marcus Rohrbach, Shih-Fu Chang, and
  Lorenzo Torresani.
\newblock Learning to recognize procedural activities with distant supervision.
\newblock In {\em IEEE Conference on Computer Vision and Pattern Recognition
  (CVPR)}, 2022.

\bibitem{liu2022video}
Ze Liu, Jia Ning, Yue Cao, Yixuan Wei, Zheng Zhang, Stephen Lin, and Han Hu.
\newblock Video swin transformer.
\newblock In {\em IEEE Conference on Computer Vision and Pattern Recognition
  (CVPR)}, 2022.

\bibitem{loshchilov2018decoupled}
Ilya Loshchilov and Frank Hutter.
\newblock Decoupled weight decay regularization.
\newblock In {\em International Conference on Learning Representations (ICLR)},
  2018.

\bibitem{miech2020end}
Antoine Miech, Jean-Baptiste Alayrac, Lucas Smaira, Ivan Laptev, Josef Sivic,
  and Andrew Zisserman.
\newblock End-to-end learning of visual representations from uncurated
  instructional videos.
\newblock In {\em IEEE Conference on Computer Vision and Pattern Recognition
  (CVPR)}, 2020.

\bibitem{miech2019howto100m}
Antoine Miech, Dimitri Zhukov, Jean-Baptiste Alayrac, Makarand Tapaswi, Ivan
  Laptev, and Josef Sivic.
\newblock {HowTo100M}: Learning a text-video embedding by watching hundred
  million narrated video clips.
\newblock In {\em IEEE International Conference on Computer Vision (ICCV)},
  2019.

\bibitem{narasimhan2022tl}
Medhini Narasimhan, Arsha Nagrani, Chen Sun, Michael Rubinstein, Trevor
  Darrell, Anna Rohrbach, and Cordelia Schmid.
\newblock {TL; DW?} summarizing instructional videos with task relevance and
  cross-modal saliency.
\newblock In {\em European Conference on Computer Vision (ECCV)}, 2022.

\bibitem{qian2022svip}
Yicheng Qian, Weixin Luo, Dongze Lian, Xu Tang, Peilin Zhao, and Shenghua Gao.
\newblock {SVIP}: Sequence verification for procedures in videos.
\newblock In {\em IEEE Conference on Computer Vision and Pattern Recognition
  (CVPR)}, 2022.

\bibitem{qiu2017learning}
Zhaofan Qiu, Ting Yao, and Tao Mei.
\newblock Learning spatio-temporal representation with pseudo-3d residual
  networks.
\newblock In {\em IEEE International Conference on Computer Vision (ICCV)},
  2017.

\bibitem{Sener_2022_CVPR}
Fadime Sener, Dibyadip Chatterjee, Daniel Shelepov, Kun He, Dipika Singhania,
  Robert Wang, and Angela Yao.
\newblock Assembly101: A large-scale multi-view video dataset for understanding
  procedural activities.
\newblock In {\em IEEE Conference on Computer Vision and Pattern Recognition
  (CVPR)}, 2022.

\bibitem{song2020mpnet}
Kaitao Song, Xu Tan, Tao Qin, Jianfeng Lu, and Tie-Yan Liu.
\newblock {MPNet}: Masked and permuted pre-training for language understanding.
\newblock {\em Advances in Neural Information Processing Systems (NeurIPS)},
  2020.

\bibitem{sun2019videobert}
Chen Sun, Austin Myers, Carl Vondrick, Kevin Murphy, and Cordelia Schmid.
\newblock {VideoBERT}: A joint model for video and language representation
  learning.
\newblock In {\em IEEE International Conference on Computer Vision (ICCV)},
  2019.

\bibitem{szegedy2016rethinking}
Christian Szegedy, Vincent Vanhoucke, Sergey Ioffe, Jon Shlens, and Zbigniew
  Wojna.
\newblock Rethinking the inception architecture for computer vision.
\newblock In {\em IEEE Conference on Computer Vision and Pattern Recognition
  (CVPR)}, 2016.

\bibitem{tang2019coin}
Yansong Tang, Dajun Ding, Yongming Rao, Yu Zheng, Danyang Zhang, Lili Zhao,
  Jiwen Lu, and Jie Zhou.
\newblock {COIN}: A large-scale dataset for comprehensive instructional video
  analysis.
\newblock In {\em IEEE Conference on Computer Vision and Pattern Recognition
  (CVPR)}, 2019.

\bibitem{tang2020comprehensive}
Yansong Tang, Jiwen Lu, and Jie Zhou.
\newblock Comprehensive instructional video analysis: The {COIN} dataset and
  performance evaluation.
\newblock {\em IEEE transactions on pattern analysis and machine intelligence},
  2020.

\bibitem{wang2022multimedia}
Qingyun Wang, Manling Li, Hou~Pong Chan, Lifu Huang, Julia Hockenmaier, Girish
  Chowdhary, and Heng Ji.
\newblock Multimedia generative script learning for task planning.
\newblock {\em arXiv:2208.12306}, 2022.

\bibitem{xu2021videoclip}
Hu Xu, Gargi Ghosh, Po-Yao Huang, Dmytro Okhonko, Armen Aghajanyan, Florian
  Metze, Luke Zettlemoyer, and Christoph Feichtenhofer.
\newblock {VideoCLIP}: Contrastive pre-training for zero-shot video-text
  understanding.
\newblock In {\em Conference on Empirical Methods in Natural Language
  Processing}, 2021.

\bibitem{xu2015discriminative}
Zhongwen Xu, Yi Yang, and Alex~G Hauptmann.
\newblock A discriminative {CNN} video representation for event detection.
\newblock In {\em IEEE Conference on Computer Vision and Pattern Recognition
  (CVPR)}, 2015.

\bibitem{zhou2018towards}
Luowei Zhou, Chenliang Xu, and Jason~J Corso.
\newblock Towards automatic learning of procedures from web instructional
  videos.
\newblock In {\em The Association for the Advancement of Artificial
  Intelligence Conference (AAAI)}, 2018.

\bibitem{zhukov2019cross}
Dimitri Zhukov, Jean-Baptiste Alayrac, Ramazan~Gokberk Cinbis, David Fouhey,
  Ivan Laptev, and Josef Sivic.
\newblock Cross-task weakly supervised learning from instructional videos.
\newblock In {\em IEEE Conference on Computer Vision and Pattern Recognition
  (CVPR)}, 2019.

\end{thebibliography}
}

\newpage
\clearpage

\setcounter{section}{0}
\setcounter{figure}{0}
\setcounter{table}{0}
\renewcommand{\thesection}{S\arabic{section}}
\renewcommand{\thetable}{T\arabic{table}}
\renewcommand{\thefigure}{F\arabic{figure}}

\section*{Supplementary Materials}

In this section, we describe additional implementation details of our method and provide more qualitative results and comparisons on all the 6 downstream tasks. 

\section{Implementation Details}
\noindent\textbf{Pre-training.} The base video model is a Timesformer~\cite{bertasius2021space} model with a ViT backbone initialized with ImageNet-21K ViT pretraining~\cite{dosovitskiy2020image}. We pre-train our model on 64 A100 GPUs for 20 epochs which takes ~120 hours for all the videos in the HowTo100M dataset. We use a batch size of 64 videos (1 video per GPU), each consisting of 12 segments. To train the model, we use SGD optimizer with momentum and weight decay. The learning rate is set to 0.01 and is decayed using a step learning rate policy of 10\% decay at steps 15 and 19. We perform a second round of pretraining for 15 epochs using AdamW~\cite{loshchilov2018decoupled} with a learning rate of 0.00005. 

We use a 15\% masking ratio during pre-training. Segment transformer $f_{\text{trans}}$ is a two layer transformer with 12 video segments as input. Each segment consists of 8 embedding vectors extracted from a series of 8 adjacent 8-second clips from the input video (spanning a total of 64 seconds). It has a 768 embedding dimnesion and 12 heads, along with learnable positional encodings at the beginning. The WikiHow knowledgebase has 10588 step classes all of which are used for training the network with step classification loss. For obtaining the distant supervision from WikiHow and mapping ASR text to step labels in the WikiHow knowledge base, we follow the setup described in ~\cite{lin2022learning}.   

\vspace{2mm}
\noindent\textbf{Fine-tuning.} For mistake step detection, mistake ordering detection, long term and short term step forecasting, and procedural activity recognition the input consists of 12 segments from the video. We fine-tune only the segment transformer $f_{\text{trans}}$ and the linear head $f_{\text{head}}$ using cross entropy loss, while the keeping the base TimeSformer video model $f_{\text{vid}}$ as a fixed feature extractor. We use a learning rate of 0.005 with a step decay of 10\% and train the network for 50 epochs using sgd optimizer.       

For the step classification task, we only fine-tune the linear head, while keeping both the base video model and the 2 layer segment transformer fixed. We use a learning rate of 0.005 with a step decay of 10\% and train the network for 50 epochs using sgd optimizer.

\section{Additional Quantitative Results}
\noindent\textbf{Activity Recognition.} In Tab.~\ref{tab:epic}, we include results for activity recognition on EPIC-KITCHENS-100 by fine-tuning our pre-trained model for noun, verb, and action recognition tasks. We outperform all baselines on noun recognition, and are on par with MoViNet~(\emph{Kondratyuk et al., CVPR 2021}) on action recognition.  

\begin{table}[h]
\centering
\footnotesize
\begin{tabular}{lllccc}
\toprule 
Model & Action (\%) & Verb (\%) & Noun (\%) 
\\ \midrule
MoViNet & \textbf{47.7} & \textbf{72.2} & 57.3  \\
LwDS: TimeSformer  & 44.4 & 67.1 & 58.1 \\ \hline
\textbf{VideoTaskformer (SC)} & \textbf{47.6} & 70.4 & \textbf{59.8} \\
\bottomrule
\end{tabular}
\caption{\textbf{Activity Recognition on EPIC-KITCHENS-100.}}
\vspace{-.1cm}
\label{tab:epic}
\vspace{-.2cm}
\end{table}

\vspace{2mm}
\noindent\textbf{Evaluating on step localization}: We evaluate our pre-trained embeddings on the action segmentation task in COIN. Following previous work, we train a linear head on top of our fixed features and predict action labels for non-overlapping 1-second 
input video segments. LwDS attains 67.6\% on this task, and our method achieves 69.1\%.

\noindent\textbf{Step labels as input}: Our method uses visual features since step labels are not always available during inference. Nevertheless, for the purpose of comparison, we assume we have access to ground-truth step labels during inference and include results for all tasks. The results shown in Tab.~\ref{tab:epic} are from training a single layer transformer on the COIN train set and evaluating on the test set, i.e. there is no pre-training. As expected, using step labels makes the task much simpler and it outperforms using visual features. However, adding task label information to visual features improves performance significantly for all the tasks.   
\begin{table}[h]
\centering
\footnotesize
\begin{tabular}{lcccccc}
\toprule 
\multirow{2}{*}{Task} & \multicolumn{2}{c}{Step Labels(\%)} & \multicolumn{2}{c}{Visual Features (\%)} \\
& - & w/ Task label & - & w/ Task label \\ \midrule
Short term forecasting & 65 & 68 & 20 & 49 \\
Long term forecasting & 50 & 53 & 14 & 40\\ 
Mistake Ordering & 80 & 82 & 60 & 65\\
Mistake Step & 64 & 68 & 28 & 33\\
\bottomrule
\end{tabular}
\caption{\textbf{Step labels vs Visual features.}}
\vspace{-.1cm}
\label{tab:epic}
\vspace{-.3cm}
\end{table}

\section{Additional Qualitative Results}
\begin{figure*}
    \centering
    \includegraphics[width=0.8\textwidth]{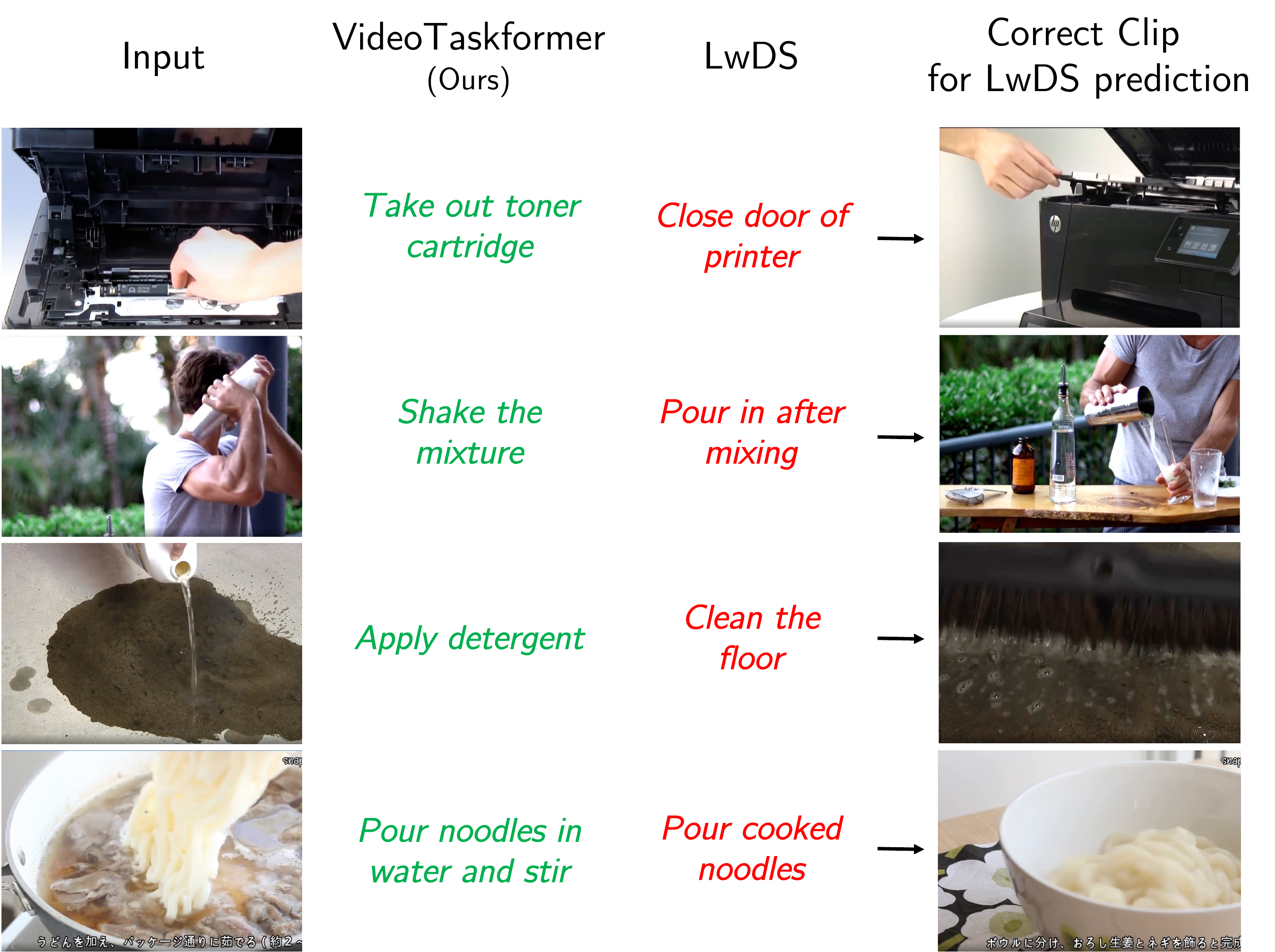}
    \caption{\textbf{Step classification.} We qualitatively compare results from our method (VideoTaskeformer) to the baseline LwDS on the step classification task. While the inputs are video clips, we only show a keyframe from the clip for visualization purposes. Correct predictions (VideoTaskformer) are shown in green and incorrect predictions (LwDS) are in red. We also show a frame from the clip corresponding to the incorrect prediction made by LwDS.}
    \label{fig:sc}
\end{figure*}

\noindent\textbf{Step Classification.} We compare results from our method VideoTaskformer, to the baseline LwDS~\cite{lin2022learning} in Fig.~\ref{fig:sc}. Since our model was trained on the entire video by masking out segments, it has a better understanding of the relationship between different steps in the same task, i.e. learned representations are \emph{``context-aware''}. As a result, it is better at distinguishing the steps within a task and correctly classifies all the steps in the four examples shown here. LwDS on the other hand incorrectly classifies all of the steps. For reference, we show a keyframe from the correct video step clip corresponding to the incorrect step class chosen by LwDS. The input image clips and the correct clips for the LwDS predictions are closely related and contain similar objects of interest, they correspond to different stages of the task and contain different actions. Since our model learns step representations \emph{``globally''} from the whole video, it is able to capture these subtle differences. 

\begin{figure*}
    \centering
    \includegraphics[width=0.9\textwidth]{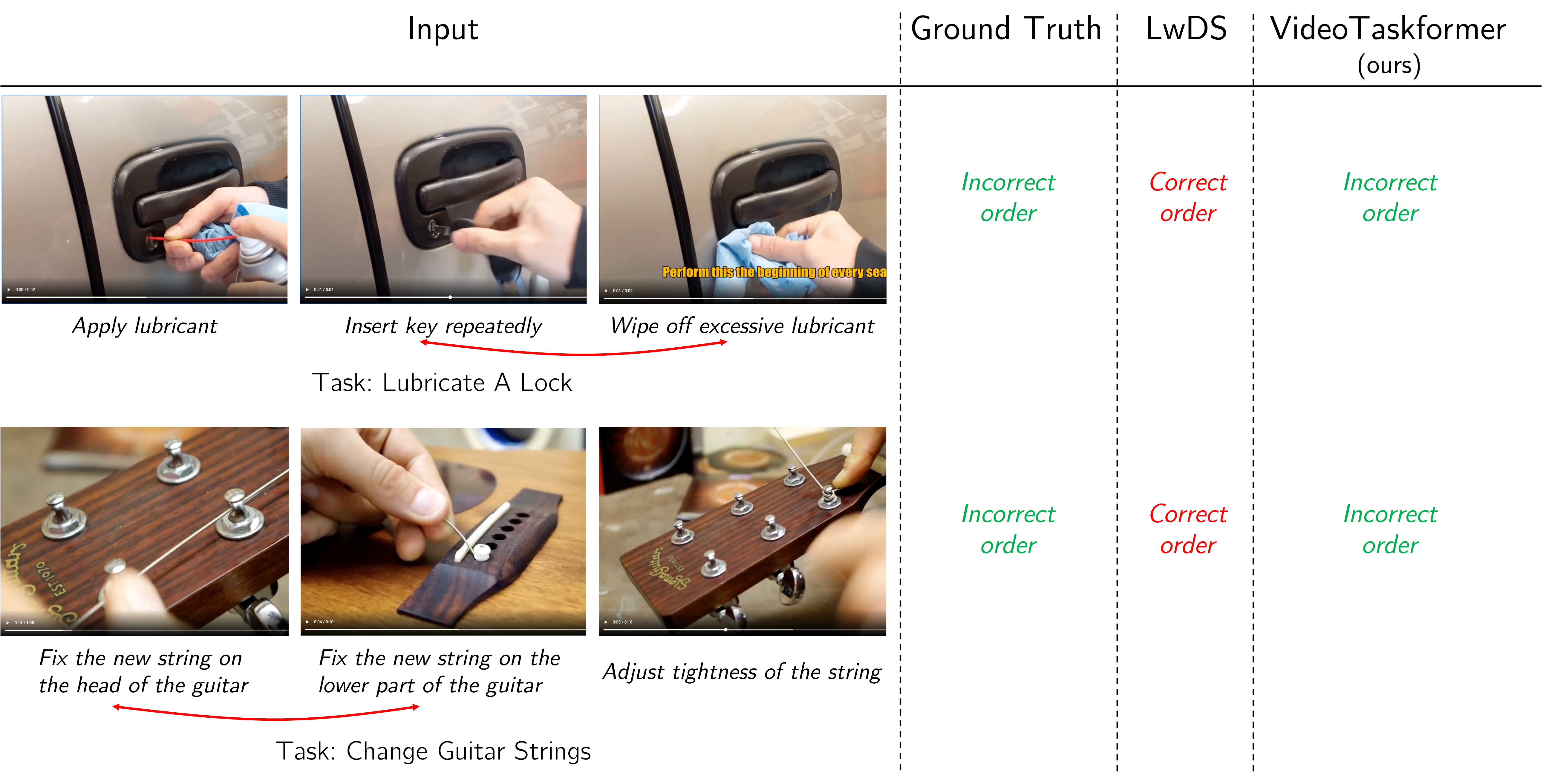}
    \caption{\textbf{Mistake Order Detection.} Qualitative comparison of results from VideoTaskformer to LwDS. Step and task labels shown along with the input are for visualization purpose only. Correct answers are shown in green and incorrect answers in red.}
    \label{fig:mo}
\end{figure*}

\begin{figure*}
    \centering
    \includegraphics[width=0.8\textwidth]{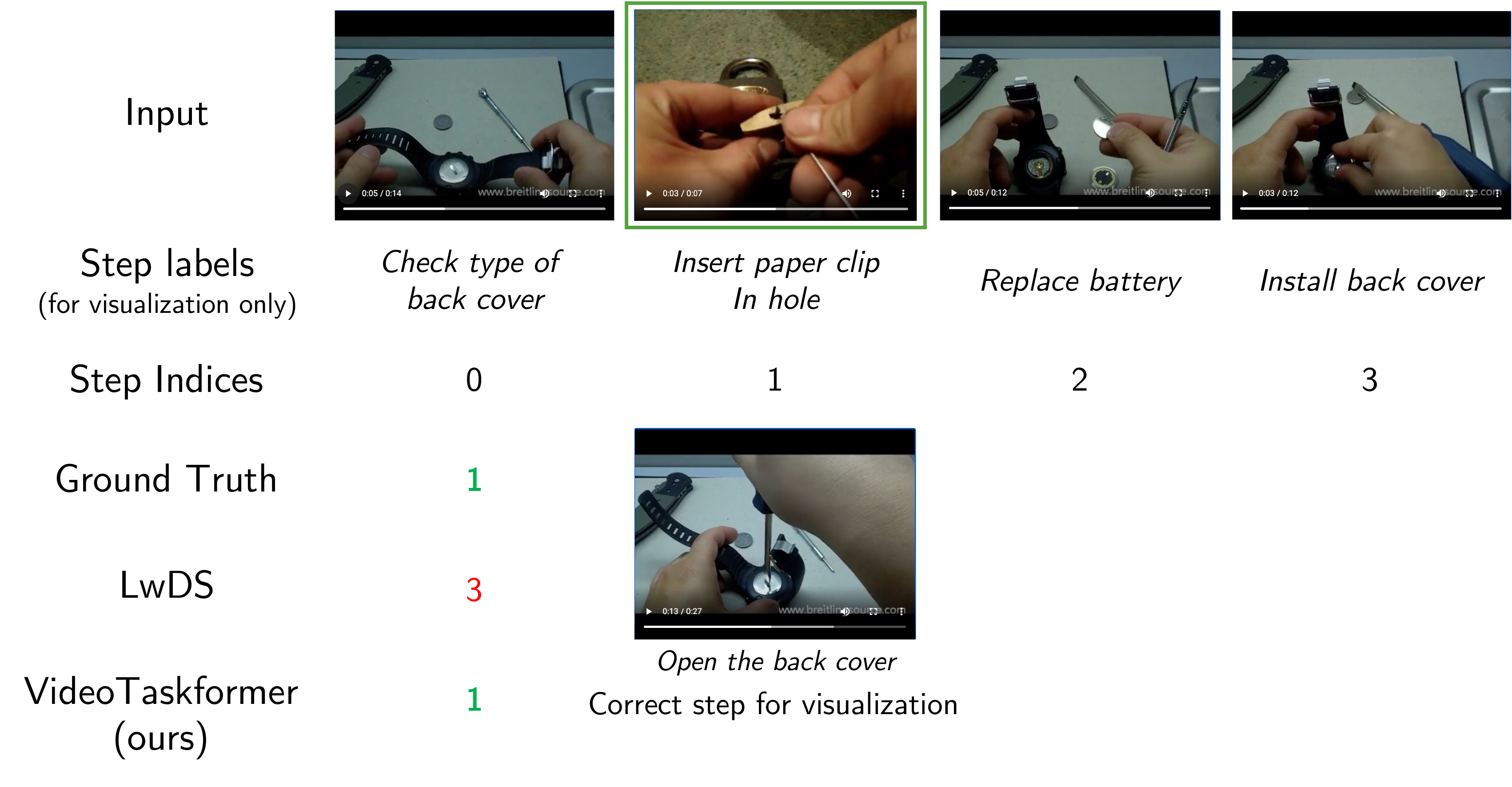}
    \caption{\textbf{Mistake Step Detection.} Qualitative comparison of results from VideoTaskformer to LwDS. Step and task labels shown along with the input are for visualization purpose only. Correct answers are shown in green and incorrect answers in red.}
    \label{fig:ms}
\end{figure*}

\begin{figure*}
    \centering
    \includegraphics[width=\textwidth]{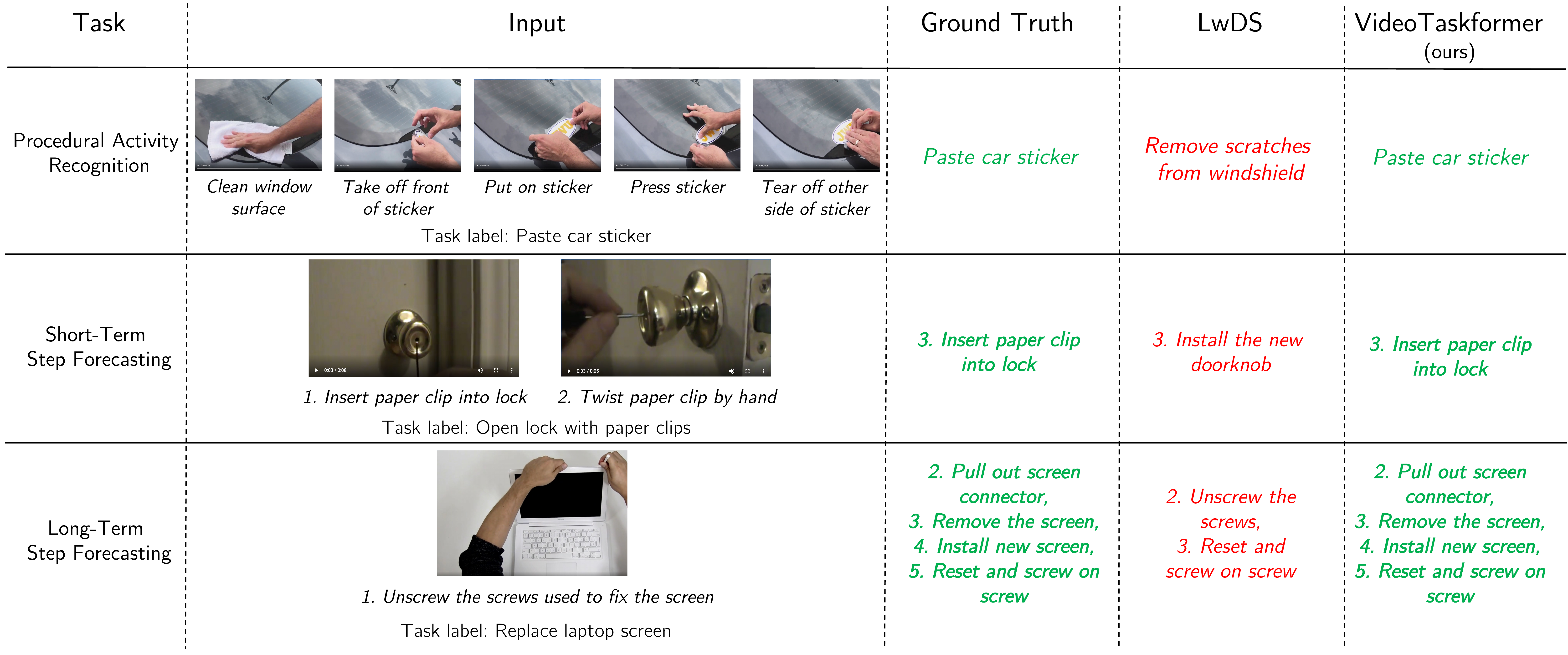}
    \caption{Qualitative results for \textbf{procedural activity recognition, short term step forecasting, and long term step forecasting}. Step and task labels shown along with the input are for visualization purpose only. Correct answers are shown in green and incorrect answers in red.}
    \label{fig:res3}
\end{figure*}

\vspace{2mm}
\noindent\textbf{Mistake Ordering Detection.} Fig.~\ref{fig:mo} compares results of our method VideoTaskformer to the baseline LwDS on the mistake ordering detection task. We show two examples, \emph{``lubricate a lock''} and \emph{``change guitar string''}, where the steps in the input are swapped as shown by red arrows. Our method correctly detects that the input steps are in the incorrect order whereas the baseline predicts the ordering to be correct. As seen, detecting the order requires a high level understanding of the task structure, which our model learns through masking. 

\vspace{2mm}
\noindent\textbf{Mistake Step Detection.} Qualitative comparison on the mistake step detection task is shown in Fig.~\ref{fig:ms}. The input consists of video clip steps for the task \emph{``change battery of watch''}. The second step is swapped with an incorrect step from a different task. Our method correctly identifies the index of the mistake step 1, whereas the baseline predicts 3 which is incorrect. We show the correct step for visualization purposes. 

\vspace{2mm}
\noindent\textbf{Procedural Activity Recognition.} A result is shown in Fig.~\ref{fig:res3}. VideoTaskformer's representations are context-aware and can identify the right task given the sequence of clips, \emph{``paste car sticker''}. The baseline misidentifies the task as an incorrect similar task, \emph{``remove scratches from windshield''}.

\vspace{2mm}
\noindent\textbf{Short-term Step Forecasting.} Fig.~\ref{fig:res3} shows an input consisting of two clips corresponding to the first two steps for the task \emph{``open lock with paper clips''}. The clips are far apart temporally, so the model needs to understand broader context of the task to predict what the next step is. Our method VideoTaskformer correctly identifies the next step as \emph{``insert paper clip into lock''} whereas the baseline incorrectly predicts a step \emph{``install the new doorknob''} from another task.

\vspace{2mm}
\noindent\textbf{Long-term Step Forecasting.} In Fig.~\ref{fig:res3} we compare the future steps predicted by our model and the baseline LwDS on the long-term step forecasting task. Both models only receive a single clip as input, corresponding to the first step \emph{``unscrew the screws used to fix the screen''} of the task \emph{``replace laptop screen''}. Our model predicts all the next 4 ground-truth steps correctly, and in the right order. The baseline on the other hand predicts steps from the same task but in the incorrect order. 

All of the above qualitative results further support the effectiveness of learning step representations through masking, and show that our learned step representations are \emph{``context-aware''} and possess \emph{``global''} knowledge of task-structure.  

%%%%%%%%% REFERENCES
% {\small
% \bibliographystyle{ieee_fullname}
% \bibliography{egbib}
% }

% \end{document}

\end{document}